# Optimization and Design of a Laser-Cutting Machine using Delta Robot

[1]B.Moharana, [2]Rakesh Gupta, [3]Bashishth Kumar Kushwaha

[1]DR. B.B.A GOVT.POLYTECHNIC, Karad, India
[2,3]Roorkee Engineering & Management Technology Institute, Shamli, India

**Abstract: -**

Industrial high-speed laser operations the use of delta parallel robots potentially offers many benefits, due to their structural stiffness and limited moving masses. This paper deals with a particular Delta, developed for high-speed laser cutting. Parallel delta robot has numerous advantages in comparison with serial robots: Higher stiffness, and connected with that a lower mass of links, the possibility of transporting heavier loads, and higher accuracy. The main drawback is, however, a smaller workspace. Hence, there exists an interest for the research concerning the workspace of robots.in industrial cutting tool maximum do not have more prescribe measurement to cut so that in This paper is oriented to parallel kinematic robots definition, description of their specific application of laser cutting, comparison of robots made by different producers and determination of velocity and acceleration parameters, kinematic analysis – inverse and forward kinematic. It brings information about development of Delta robot. The production of laser cutting machines began thirty years ago. The progress was very fast and at present time every year over 3000 laser cutting machines is installed in the world. Laser cutting is one of the largest applications of lasers in metal working industry.

## I. INTRODUCTION

Parallel architectures have the end-effector (platform) connected to the frame (base) through a number of kinematic chains (legs). Their kinematic analysis is often difficult to address. The analysis of this type of mechanisms has been the focus of much recent research. Stewart presented his platform in 1965 [1]. Since then, several authors [2], [3] have proposed a large variety of designs. The interest for parallel manipulators robot (PMR) arises from the fact that they exhibit high stiffness in nearly all configurations and a high dynamic performance. Recently, there is a growing tendency to focus on parallel manipulators with 3 translational [4, 5,]. In the case of the three translational parallel manipulators, the mobile platform can only translate with respect to the base. The DELTA robot (see figure 1) is one of the most famous translational parallel manipulators robot [5, 6, 7, 11, and 12].

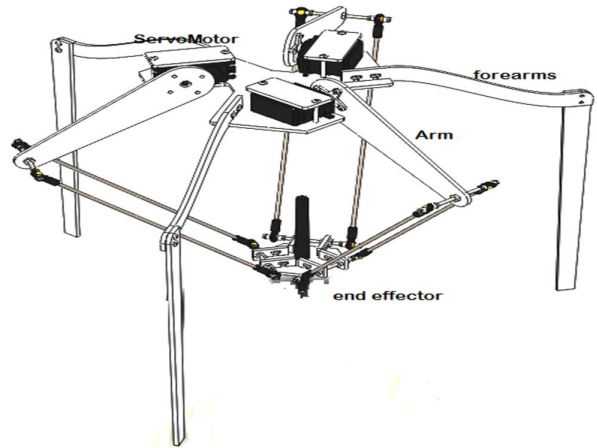

Figure 1 Delta robot sketch model with three degree angle freedom

Laser beam device should connect to mechanical systems that allow a rigid body (called end effector) to move with respect to a fixed base play a very important role in numerous applications. A rigid body in space can move in various ways, in translation or rotary motion. These are called its degrees of freedom (DOF). The position and the orientation of the end-effector (called its pose) can be described by its generalized coordinates. As soon as it is possible to control several degrees of freedom of the end-effector via a mechanical system, this system can be called robot. [4] From this aspect, robot represents an integrated Mechatronic system consisting of three subsystems: sensorial, control and decision-making, and actuation subsystem. The sensorial subsystem establishes a feedback with the environment. The control and decision-making subsystem represents the 'thinking' centre of the robot, its 'brain'. Together, the sensorial and control subsystems make up a cognitive system. Finally, the actuation subsystem is used to affect the environment, making such an impact on it that the environment changes [2]. It was concluded that the serial robots are inappropriate for tasks requiring either the manipulation of heavy loads, or a good positioning accuracy, or to work with high dynamic parameters.

## II.LASER CUTTING MACHINES

Laser machines for contour cutting thin sheet present the product of high technology. They are composed of: laser, beam guiding, cutting head, coordinate table, system for energy supply and control unit. It is based on vaporize the material in a very small area by focused laser beam. Process characteristics are: uses a high energy beam of coherent light; beam is focused on a small spot on the work piece by a lens; focused beam melts, vaporizes, or combusts material; molten material is ejected out from the melt area by pressurized gas jet. Laser cutting is the high speed cutting with a narrow kerf width that results in superior and enhanced quality, higher accuracy and greater flexibility. The optical quantum generator generates the light beam that presents a tool in working process. By optical system in cutting head the laser beam is focused in diameter from 0.2 mm with the power density over $10^8$ W/cm2. Since our desire is to remove the evaporated and molten material from the affected zone as soon as possible, the laser cutting is performed with a coaxial assist gas. The gas blowing increases the feed rate for as much as 40 %. Cutting process along contour is realized with the movement of laser beam or work piece. Machine for movements is accordance with necessities of laser machine. [14] Laser cutting robots have totally flying optics architecture. All movements are made by the focusing head and the work-piece remains stationary. Laser robot takes the laser beam along any continuous pre-programmed path in three-dimensional space, and then cuts with accuracy, speed and quality. The speed and acceleration on the main axes of the new generation of robotic systems are 60 m/min and 23 m/s2. Integration of the laser beam guiding in the robot arm structure offers great advantages compared to conventional systems with external laser beam guiding. In case to design more accuracy beam guiding in work space so that we are going to design some new method to implementation tool to developed in challenging to problem.

## III. CHALLENGING PROBLEM

A more challenging problem is designing a parallel manipulator delta robot for a given workspace. This problem has been addressed by Boudreau and Gosselin [5, 6], an algorithm has been worked out, allowing the determination of some parameters of the parallel manipulators using a genetic algorithm method in order to obtain a workspace as close as possible to a prescribed one. Kosinska et al. [7] presented a method for the determination of the parameters of a Delta-4 manipulator, where the prescribed workspace has been given in the form of a set of points. Snyman et al. [8] propose an algorithm for designing the planar 3-RPR manipulator parameters, for a prescribed (2-D) physically reachable output workspace. Similarly in [9] the synthesis of 3-dof planar manipulators with prismatic joints is performed using GA, where the architecture of a manipulator and its position and orientation with respect to the prescribed workspace were determined.

### a. KINEMATIC ANALYSIS

If we want to build our own Delta robot, we need to solve two problems. First, if we know the desired position of the end effector (for example, we want to catch some object in the point with coordinates X,Y, Z), we need to determine the corresponding angles of each of the three arms (joint angles) to set motors (and, thereby, the end effector) in proper position for picking. The process of such determining is known as inverse kinematics.

And, in the second case, if we know joint angles (for example, we have read the values of motor encoders), we need to determine the position of the end effector (e.g. to make some corrections of its current position). This is a forward kinematics problem. [10]

To be more formal, we can look at the kinematic scheme of a delta robot. The platforms are two equilateral triangles: the fixed one with motors is green, and the moving one with the end effector is pink. Joint angles are $\theta_1$, $\theta_2$ and $\theta_3$ and point E is the end effector position with coordinates $(x_0, y_0, z_0 )$. To solve inverse kinematics problem we have to create the function with E coordinates $(x_0, y_0, z_0)$ as parameters which returns $(\theta_1, \theta_2, \theta_3)$ Forward kinematics functions gets $(\theta_1, \theta_2, \theta_3)$ and returns $(x_0, y_0, z_0)$.

### b. INVERSE KINEMATICS

The Delta robot consists of a moving platform connected to a fixed base through three parallel kinematic chains. Each chain contains a rotational joint activated by actuators in the First; let's determine some key parameters of the robot's geometry. Let's designate the side of the fixed triangle as f, the side of the end effector triangle as e, the length of the upper joint as $r_f$ , and the length of the parallelogram joint as $r_e$. These are physical parameters which are determined by the design of Delta robot. The reference frame will be chosen with the origin at the centre of symmetry of the fixed triangle, as shown below, so z -coordinate of the end effector will always be negative.

Because of the robot's design the joint $F_1$, $J_1$ (Fig. 3) can only rotate in YZ plane, forming a circle with the centre in point $F_1$ and radius $r_f$. As opposed to $F_1$, $J_1$ and $E_1$ are so-called universal joints, which means that $E_1J_1$ can rotate freely relatively to $E_1$ , forming a sphere with the centre in point $E_i$ and radius $r_f$ .

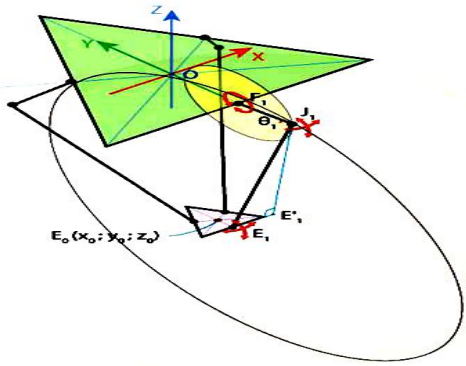

Figure 2 Delta robot joint moving parameters sketch

Intersection of this sphere and YZ plane is a circle with the centre in point $E_1'$ and radius $E_1'$ $J_1$, where $E_1'$ is the projection of the point $E_1$ on YZ plane. The point $J_1$ can be found now as intersection of two circles of known radius with centres in $E_1'$ and $F_1$ intersection point with smaller -coordinate). And if we know $J_1$, we can calculate angle $\theta_1$ [10].

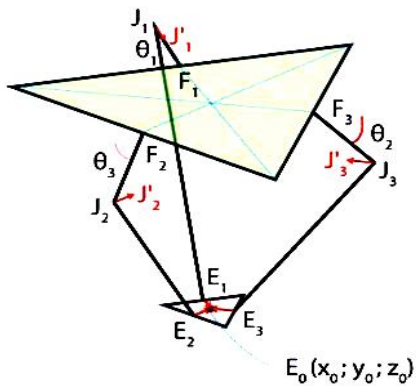

Figure 3 coordinates of end effector ($E_0$ calculation)

#### c. FORWARD KINEMATICS

In this case the three joint angles $\theta_1$, $\theta_2$, $\theta_3$ are given, and we need to find the coordinates $(x_0, y_0, z_0)$ of end effector point $E_1$ . As we know angles theta1, we can easily find coordinates of points $J_1$, $J_2$ and $J_3$ (Fig. 3). Joints $J_1E_1$, $J_2E_2$ and $J_3E_3$ can freely rotate around points $J_1$, $J_2$ and $J_3$ respectively, forming three spheres with radius $r_e$.Now let's do the following: move the centres of the spheres from points $J_1$, $J_2$ and $J_3$ to the points $J_1'$, $J_2'$ and $J_3'$ using transition vectors $E_1E_0$, $E_2E_0$ and $E_3E_0$ respectively. After this transition all three spheres will intersect in one point $E_0$.So, to find coordinates $(x_0, y_0, z_0)$ of point $E_0$, we need to solve the set of three equations like $(x - x_j)^2 + (y - y_j)^2 + (z - z_j)^2 = r_e^2$ , where coordinates of sphere centres $(x_j, y_j, z_j)$ and radius $r_e$ are known.

#### IV.DESIGN A CONTROL SYSTEM

For the control of the machine the commercial motion controller ROBOX RBXE has been chosen (up to 3 axes, Intel Atom microprocessor, 1.2GHz, endowed with teach pendant). It offers the user the possibility to implement her/his own defined model based control laws and coordinates transformation between the world Cartesian coordinate system and the robot internal coordinate system minimizing the effort of writing interpolating algorithms [13][15]. All the control algorithms are written in C++ and the control state variables' references updating is made every 2.5 milliseconds. The programming environment runs on Windows NT based personal computers and, besides allowing to write, compile and debug the application software, it permits to evaluate the behaviour of the controlled machine and then to choose the best solution. Particular attention has been paid to the robot working safety. Special safety measures have been integrated into the control system including digital watch dog modules with the aim to assure intrinsic safety (Fig. 4).

The safety watchdog process in the design's safety layer provides an overall health check on the system. It performs two major functions: resetting the hardware watchdog timer and monitoring to ensure that other software processes are still running. At regular intervals, all processes are required to send a pulse to the safety watchdog process: failure in receiving the signal within the specified time period causes a corrective action to occur. Corrective actions may vary depending on which process has failed and the severity of the failure.

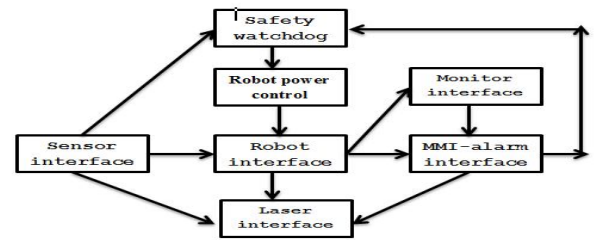

**Figure 4** design of the robot control system

#### V. CONCLUSION

The overall performance and industrial feasibility of a laser cutting system with parallel kinematics structure have been assessed. The limited moving masses that characterise this type of mechanical system allow obtaining high speed and acceleration with a relatively low power consumption. The robot behaviour can be implemented on a commercial motion controller for robotic applications, the past, one of the main hindrances to the diffusion of parallel robots was the computational heaviness of kinematics; this problem has been overcome thanks to the impressive progress in computer technology of the last years.

direct and inverse kinematic equations that describe the

## ABOUT THE AUTHORS

**Bhimsen Moharana** received the B.Tech degree in

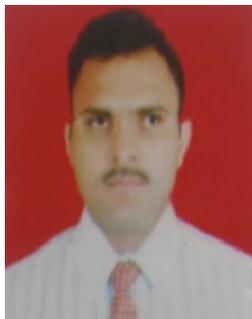

Mechanical Engineering from Utkal University, Udisha, India in 1990. He Persuining M.Tech degree from NITTIR, Chandigarh, India.During his graduation work, he esearched on various, Robotics, tools designing, materials. He is currently working as Lecturer in DR. B.B.A GOVT.POLYTECHNIC, Karad, India.

**Rakesh Gupta** received the B.E. Degree in

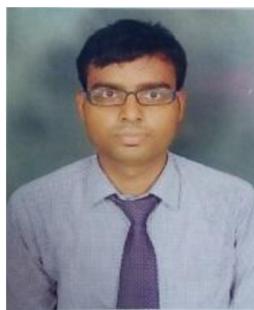

Electrical and Electronic Engineering from Anna University, Chennai, India. During his Graduation he held several workshop and summer training program. He is currently working as Assistant Professor in Roorkee Engineering And Management Technology Institute, Uttar Pradesh, India. He has research interest in low power system design, Robotic, Embedded system, VHDL.

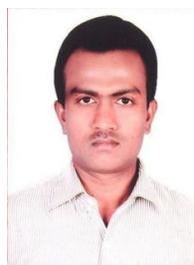

**Bashishth Kumar Kushwaha** born in deoria,India.He received the B.Tech degree in mechanical engineering from United College of Engineering and Research nainee, Allahabad, India in 2008.During his Graduate Studies time he held many research training program. He is currently working as Assistant Professor in Roorkee Engineering And Management Technology Institute, Uttar Pradesh, India. He has research interest in Material, Tool cutting and robotics.